\pdfoutput=1
\documentclass[11pt]{article}
\usepackage[table,xcdraw]{xcolor}
\usepackage[final]{acl}
\usepackage{amssymb}
\usepackage{color}
\usepackage{colortbl}
\usepackage{tcolorbox}
\usepackage{multirow}
\usepackage{times}
\usepackage{latexsym}

\usepackage{amsmath}
\usepackage[T1]{fontenc}
\usepackage{float}

\usepackage[utf8]{inputenc}

\usepackage{microtype}

\usepackage{inconsolata}

\usepackage{graphicx}

\title{Plug-and-Play Training Framework for Preference Optimization}

\author{
Jingyuan Ma$^1$ , Rui Li$^1$, Zheng Li$^1$ , Lei Sha$^2$ , Zhifang Sui$^1$
\\
$^1$School of Computer Science, Peking University\\
$^2$Institute of Artificial Intelligence, Beihang University\\
{\texttt{{\{mjy}@stu.pku.edu.cn\}}}
}
\begin{document}
\maketitle
\begin{abstract}

Recently, preference optimization methods such as DPO have significantly enhanced large language models (LLMs) in wide tasks including dialogue and question-answering. 
However, current methods fail to account for the varying difficulty levels of training samples during preference optimization, leading to mediocre performance in tasks with high accuracy requirements, particularly in mathematical reasoning.
To address this limitation, we propose a novel training framework, which employs multiple sampling to analyze output distributions, assign different weights to samples, and incorporate these weights into the preference optimization process. 
This plug-and-play approach enables LLMs to prioritize challenging examples during training, improving learning efficiency.
Experimental results demonstrate that our framework integrates seamlessly with various preference optimization methods and achieves consistent improvements in mathematical reasoning tasks.

\end{abstract}

\section{Introduction}

Preference Optimization (PO) methods, such as Direct Preference Optimization (DPO)~\citep{dpo} and Proximal Policy Optimization (PPO)~\citep{ppo}, have become widely used techniques for aligning large language models (LLMs) with human preferences. These methods have achieved remarkable success across tasks such as dialogue generation and question answering, where alignment with human-like outputs is critical~\citep{rlhf}. 
\begin{figure}[!t]
    \centering
    \includegraphics[width=1\linewidth]{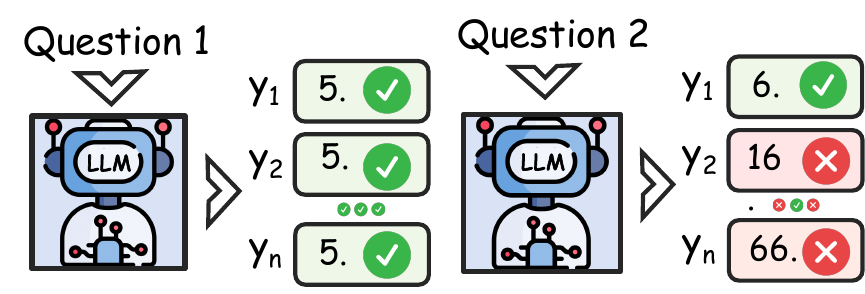}
    \caption{The figure illustrates the variability in the model's output when sampling the same question multiple times. In \textbf{Question 1}, the model consistently produces the correct answer across all samples. In contrast, \textbf{Question 2} demonstrates a case where the model generates diverse responses, including incorrect answers, reflecting uncertainty or inconsistency in the model's reasoning process.}
    \label{fig:different}
\end{figure}

However, a fundamental limitation of existing PO methods is their inability to dynamically account for the varying difficulty levels of training samples, which refers to the model's degree of mastery over different samples. This uniform weighting approach assumes that all samples contribute equally to the learning process, overlooking critical differences in sample complexity and model error tendencies.
For tasks requiring high accuracy, such as mathematical reasoning, this issue becomes particularly pronounced. 

Our initial analysis examines the output distributions of LLMs when presented with mathematical problems of varying difficulty.
As illustrated in Figure~\ref{fig:different} and further analyzed in Section~\ref{sec:tiny}, the output distribution can be a model-specific indicator of the relative difficulty of the problem to the LLM.
For example, the model consistently generates correct answers towards simple questions, indicating a strong understanding of the question. However, in more challenging cases, the model produces diverse and often incorrect answers, suggesting an unstable grasp of the problem. 
Furthermore, the distribution of specific error types in the model's outputs highlights the most detrimental ``bad'' responses, which should have their probabilities reduced to improve performance.

Leveraging these insights,
we propose a novel \textbf{plug-and-play weighted training framework}, which can be seamlessly integrated into various Preference Optimization methods. The framework consists of three key stages: a data collection phase, a weighted optimization phase and weighted training phase. In the first stage, inspired by \citet{monkeys} and \citet{testtimesample}, training data is collected by sampling the model's responses multiple times for the same question. This approach provides a better understanding of the model's output distribution and preferences. In the second stage, we design a metric to adjust the weight of each training sample according to the model's performance on each question. By analyzing the frequency of correct and incorrect responses through repeated sampling, we assigns higher weights to challenging samples where the model struggles, while reducing the emphasis on samples that the model has already mastered. In the third stage, the training process that leverages the weights we get to prioritize challenging samples, effectively optimizing the model's focus during training.
This approach can be applied to various Preference Optimization methods, such as DPO~\citep{dpo}, DPOP~\citep{dpop} and IPO~\citep{ipo}, or other pairwise preference-based alignment techniques, providing a flexible and effective way to improve the utilization of training data. By aligning training more closely with the model's output patterns and focusing on difficult examples, this adaptable framework enhances the efficiency of training and drives better model performance across diverse optimization tasks.
Our contributions are summarized as follows:
\begin{itemize}
\item We propose a novel framework that introduces a data collection process and a metric for computing dynamic weights based on the model's output distribution.
\item We demonstrate that these weights can be seamlessly integrated into a variety of pairwise comparison preference optimization algorithms.
\item Experimental results on multiple model series show that our framework effectively improves the mathematical reasoning capabilities of LLMs.
\end{itemize}

\section{Related Work}

\paragraph{Preference Optimization}
Preference Optimization (PO) methods have become a cornerstone for aligning large language models (LLMs) with human preferences, particularly within Reinforcement Learning from Human Feedback (RLHF) frameworks~\citep{rlhf}. These methods leverage reward models to score outputs based on human feedback and guide the model to generate preferred responses. For instance, Proximal Policy Optimization (PPO)~\citep{ppo} is a widely adopted approach that uses reward signals to optimize the likelihood of preferred responses while maintaining output diversity. While these reward-based methods have shown success in general alignment tasks such as dialogue generation~\citep{alpaca_eval}, recent studies have observed performance degradation in more structured and high-precision tasks, such as mathematical reasoning~\citep{simpo, DBLP:conf/icml/LiuGBCBLHDVB24, dpop}. Specifically, such methods often overemphasize stylistic alignment rather than reasoning accuracy, leading to suboptimal outcomes in domains where correctness is paramount. 

\paragraph{Pairwise Comparison Optimization}
Among Preference Optimization approaches, pairwise comparison methods are particularly notable for their simplicity and effectiveness. These methods learn by comparing pairs of outputs, typically labeled as ``chosen'' (preferred) and ``rejected'' (non-preferred). Direct Preference Optimization (DPO)~\citep{dpo} is a widely recognized algorithm in this category, which directly optimizes the model to increase the probability of preferred responses. However, DPO can suffer from overfitting, particularly when handling limited or imbalanced preference data. To address this, IPO~\citep{ipo} was introduced, which refines the training process to improve generalization performance. Additionally, SimPO~\citep{simpo} simplifies pairwise comparison-based methods by removing the reference model dependency, reducing both computational costs and training complexity. DPOP (DPO-positive)~\citep{dpop} further improves robustness by addressing challenges such as minimal edit distance between chosen and rejected responses, which can mislead the optimization process. These methods collectively highlight the potential of pairwise comparison-based approaches to deliver efficient and scalable alignment solutions, particularly in preference-driven tasks.

\section{Plug-and-play Weighted Training}

\begin{figure*}[!t]
    \centering
    \includegraphics[width=1\linewidth ]{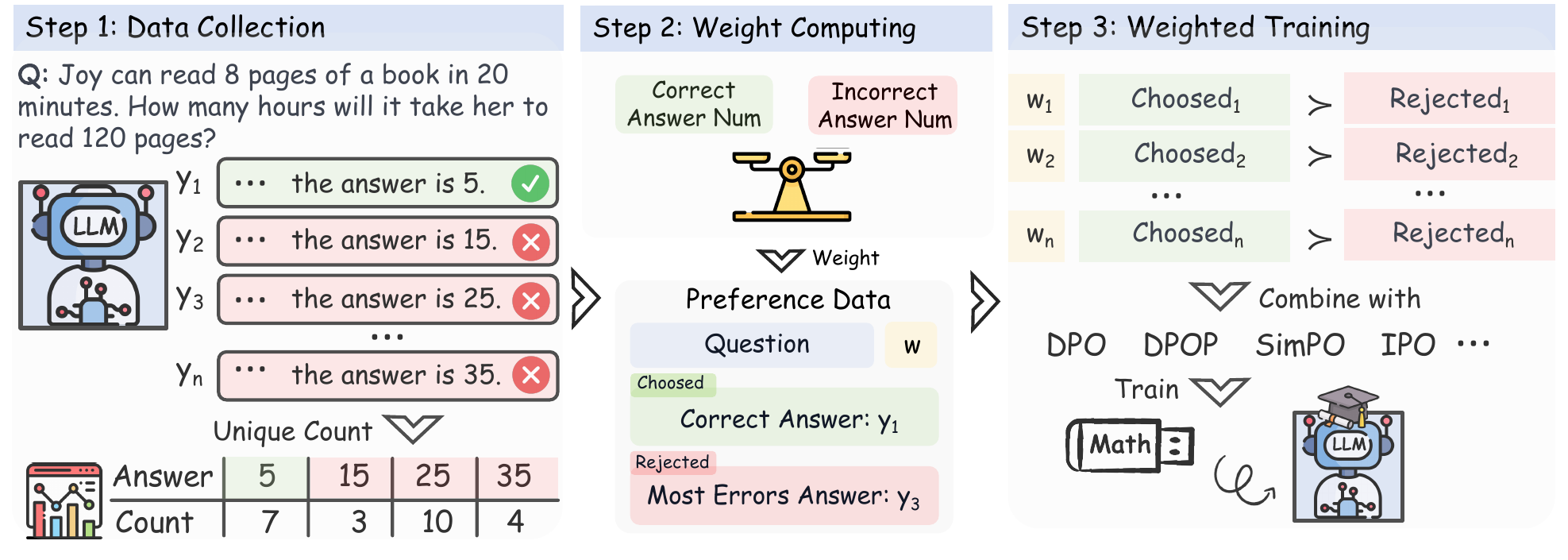}
    \caption{Overall process of the framework. \textbf{In Step 1}, we begin by sampling the model multiple times to collect the distribution of responses for each question. \textbf{In step 2}, we identify the responses with the highest number of incorrect answers as well as the correct answers. These are then weighted according to their frequency of occurrence. \textbf{In step 3}, various pairwise comparison alignment methods can be applied to incorporate these weights into the training process, ultimately resulting in a trained model.
}
    \label{fig:mainplot}
\end{figure*}

Training large language models often encounters the challenge of imbalanced data utility—some examples are well-understood by the model, while others remain poorly mastered. Traditional training methods treat all data equally, which can lead to overfitting on simpler examples while underutilizing more challenging ones. Our weighted training framework addresses this issue by computing the importance of training examples based on the model's performance. As illustrated in Figure~\ref{fig:mainplot}, we leverage multiple sampling to capture the model's output distribution and tendencies, using this information to assign a weight to each data pair. These weights are then integrated into existing training methods, allowing the model to focus more effectively on challenging examples.

\subsection{Understanding Sampling Behavior and Answer Distribution}
\label{sec:tiny}
\begin{figure}[t!]
    \centering
    \includegraphics[width=1\linewidth]{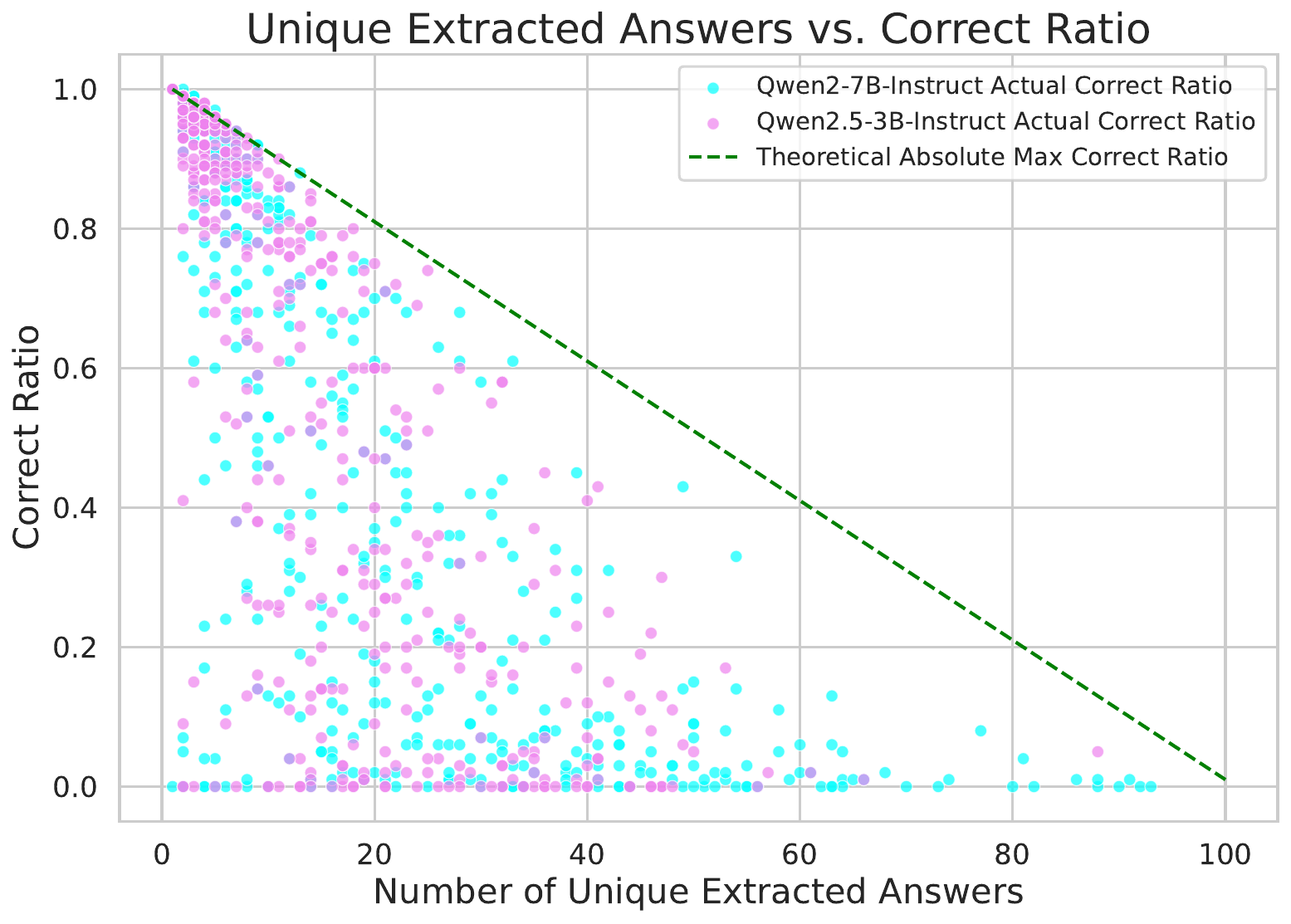}
    \caption{The distribution of the responses of the model in the face of multiple samples of the same question. Each point represents a question. Here, the x-axis represents the number of different unique answers obtained in the sampling(which can be viewed as the number of answer equivalence classes), and the y-axis represents the proportion of correct answers among all 100 responds.}
    \label{fig:correct_ration}
\end{figure}

We first conduct a preliminary experiment in which we sample each question 100 times using the same model. We then analyze the number of unique answers (i.e., the number of distinct numeric answers the model generated among the 100 samples) and the proportion of correct answers among them. The results are presented in Figure~\ref{fig:correct_ration}. Additionally, we calculate the theoretical maximum accuracy under an extreme assumption. This defines the theoretical upper bound of accuracy based on the number of unique answers $k$ and the total number of samples $N$, assuming that the correct answer is the most frequently generated response. The theoretical accuracy is expressed as:

\begin{equation}
    \text{Acc}_{\text{max}} = \frac{N - (k - 1)}{N}
\end{equation}
where \( N \) is the total number of samples. \( k \) denotes the number of unique extracted answers. When \( k = 1 \), the theoretical maximum accuracy is \( 1.0 \), meaning all generated answers are correct. Conversely, when \( k = N \), the theoretical maximum accuracy approaches \( \frac{1}{N} \), indicating that the generated answers are entirely dispersed with no central tendency.

We find that many data points are concentrated in the upper left corner, indicating that the model provides relatively consistent answers for certain questions, often closer to the correct answer. However, the data points in the bottom left suggest that the model frequently generates the same incorrect answer for some questions. These are the areas our method aims to address, with the goal of shifting these data points upward through training.

\subsection{Data Collection and Weight Computing}
\label{sec:datacollection}
To better evaluate whether the model has mastered a given math problem, we use the same prompt to sample the problem multiple times, forming a dataset \( D = \{x, y_1, y_2, \ldots, y_n\} \), where \( x \) is the question and \( y_1, y_2, \ldots, y_n \) are the corresponding responses. We then process these \( n \) responses to extract their answers \( \{a_1, a_2, \ldots, a_n\} \). Next, we compute the number of correct responses \( P_{\text{c}} \), and the number of incorrect responses \( P_{\text{e}} \). Using these values, we assign weights to each question-answer pair based on the following formula:
\begin{equation}
\label{eq:getdata}
w =
\begin{cases}
1 + \alpha \cdot \frac{P_{\text{e}}}{\text{N}}, & \text{if } P_{\text{c}} = 0, \\
\max\left(1, 1 + \alpha \cdot \frac{P_{\text{e}}}{P_{\text{c}} + \epsilon} \cdot \frac{1}{\text{N}}\right), & \text{if } P_{\text{c}} > 0.
\end{cases}
\end{equation}
where \( \alpha \) is a hyperparameter that controls the magnitude of weight adjustment, while \( \epsilon \) is a small constant. The parameter \( \text{N} \) represents the maximum number of sampling attempts for each question. Additionally, the \( \max(1, \cdot) \) function ensures that the weight \( w \) is at least 1, preventing the data's importance from being excessively reduced.
An intuitive understanding of this weighting scheme is that when the model encounters a problem and consistently makes the same mistake (i.e., it repeatedly generates the same incorrect answer), this indicates that the model has developed a systematic misunderstanding of the problem. In such cases, a relatively large penalty is needed, which is reflected in a higher weight. Conversely, if the model answers most of the samples correctly and only makes occasional mistakes, it suggests that the model has already mastered this problem. Consequently, this set of data requires less emphasis in training (i.e., fewer weight adjustments), as the model does not need significant corrections.

Next, we construct the preference data pair \( D = \{x, y_w, y_l\} \), where \( y_w \) represents the model's correct answer along with its associated reasoning process, selected as the chosen answer, and \( y_l \) is the response with the most errors, designated as the rejected answer. If the model fails to generate a correct answer, we default to using the gold answer as the chosen answer. We select the model's own response as the chosen answer because while the gold answer may appear more standardized, the model's response aligns better with its own distribution. This alignment helps avoid forcing the model to mimic the gold answer's style, allowing it to focus on learning effective reasoning steps. After all, in the mathematical domain, human preference and writing style is secondary to following the correct steps to arrive at the correct answer.

This weighting mechanism ensures that each question-answer pair is appropriately weighted based on the model's performance, allowing the framework to emphasize challenging problems and better utilize training data.

\subsection{Weighted Training}
\label{sec:weightedmethod}
In this section, we explain how the weights obtained in the previous step are used during training. When we have two responses, \( y_w \) and \( y_l \)\footnote{Where w represents win and l represents loss.}, the Bradley-Terry model~\citep{btmodel} assumes that the probability of the model outputting \( y_w \) is greater than the probability of outputting \( y_l \). The overall optimization objective of the reward function can therefore be designed as:
\begin{equation}
L_R = -\mathbb{E}_{(x, y_w, y_l) \sim D} \left[ \log P(y_w \succ y_l | x) \right]
\end{equation}

We denote \( r(x, y_w) \) as the score of the response \( y_w \), so that the difference \( r(x, y_w) - r(x, y_l) \) represents the model's internal scoring of these two responses. This step can be understood as evaluating the relative scores of the two responses from the model's internal perspective, since it reflects the distribution of the model's output logits.

Additionally, the external ``score'' represented by the weight \( \textbf{w} \), which we get from the equation \ref{eq:getdata}, quantifies the importance of the pair for model optimization. This leads to the updated objective:
\begin{multline}
    L_R = -\mathbb{E}_{(x, y_w, y_l) \sim D} [ 
    \log \sigma \big( \textbf{w} \cdot  (r(x, y_w) \\
    - r(x, y_l)) \big) ]
\end{multline}
Since the optimal reward can be represented as the optimal value for the model\citep{peng2019advantage, peters2007reinforcement}, we can obtain:
\begin{equation}
    r(x, y) = \beta \log \frac{\pi^*(y | x)}{\pi_{\text{ref}}(y | x)} + \beta \log Z(x)
\end{equation}
where $\pi^*$ is the optimal model and the $\text{Z(x)}=\sum_y\pi_{ref}(y|x)exp(\frac1\beta r(x,y))$ is a partition function of $x$ so that it is unrelated to the model we are trying to optimize. So substituting the equation into the objective, the core part would be:
\begin{equation}
    \textbf{w} \Big( 
      \log \frac{\pi^*(y_w | x)}{\pi_{\text{ref}}(y_w | x)} \\
    -  \log \frac{\pi^*(y_l | x)}{\pi_{\text{ref}}(y_l | x)}
    \Big) 
\end{equation}
Unlike other pairwise comparison methods, which may enhance the alignment process through techniques such as length penalty control or reference-model-free approaches, our method focuses on optimizing the utilization of reward differences to improve training process. In practice, this computation can be implemented efficiently by incorporating \( w \) into the log-probability calculation, adding negligible computational overhead.

\section{Experiments}

\begin{table*}[t!]
    \centering
    \caption{The main experimental results. It can be observed that the performance of models trained using the weighted data method on our collected datasets improved across several methods. \textbf{All the models were trained on the MetaMath-MATH dataset (not include GSM8K data).} MATH500 can be considered as in-domain data. Additionally, we also present the performance of each model on the corresponding dataset without additional training for comparison.}
    \label{tab:results}
    \begin{tabular}{p{4.5cm} >{}p{4cm} >{}p{2cm} >{\columncolor{gray!15}}p{2cm}}
        \hline
        \textbf{Model} & \textbf{Method} & \textbf{GSM8K} & \textbf{Math500} \\
        \hline
        \multirow{9}{*}{\centering Qwen2-1.5B-Instruct} & Original & 63.22 & 22.8 \\
        & SFT & 65.42 {\scriptsize (\textit{+2.20})} & 25.4 {\scriptsize (\textit{+2.6})} \\
        & DPO & 64.74 {\scriptsize (\textit{+1.52})} & 24.8 {\scriptsize (\textit{+2.0})} \\
        & \quad \textit{with weight data} & 64.59 {\scriptsize (\textit{+1.37})} & 25.6 {\scriptsize (\textit{+2.8})} \\
        & DPOP & 64.74 {\scriptsize (\textit{+1.52})} & 24.4 {\scriptsize (\textit{+1.6})} \\
        & \quad \textit{with weight data} & 64.36 {\scriptsize (\textit{+1.14})} & \textbf{26.0 {\scriptsize (\textit{+3.2})}} \\
        & SimPO & 64.74 {\scriptsize (\textit{+1.52})} & 25.6 {\scriptsize (\textit{+2.8})} \\
        & \quad \textit{with weight data} & 65.42 {\scriptsize (\textit{+2.20})} & 25.9 {\scriptsize (\textit{+3.1})} \\
        & IPO & 64.74 {\scriptsize (\textit{+1.52})} & 25.8 {\scriptsize (\textit{+3.0})} \\
        & \quad \textit{with weight data} & 64.74 {\scriptsize (\textit{+1.52})} & 25.2 {\scriptsize (\textit{+2.4})} \\
        \hline
        \multirow{9}{*}{\centering Qwen2-7B-Instruct} & Original & 84.68  & 51.0 \\
        & SFT & 84.98 {\scriptsize (\textit{+0.30})} & 53.8 {\scriptsize (\textit{+2.8})} \\
        & DPO & 87.49 {\scriptsize (\textit{+2.81})} & 55.8 {\scriptsize (\textit{+4.8})} \\
        & \quad \textit{with weight data} & 87.49 {\scriptsize (\textit{+2.81})} & \textbf{57.6 {\scriptsize (\textit{+6.6})}} \\
        & DPOP & 84.76 {\scriptsize (\textit{+0.08})} & 53.2 {\scriptsize (\textit{+2.2})} \\
        & \quad \textit{with weight data} & 85.44 {\scriptsize (\textit{+0.76})} & 53.2 {\scriptsize (\textit{+2.2})} \\
        & SimPO & 85.82 {\scriptsize (\textit{+1.14})}  & 52.6 {\scriptsize (\textit{+1.6})} \\
        & \quad \textit{with weight data} & 85.36 {\scriptsize (\textit{+0.68})} & 53.4 {\scriptsize (\textit{+2.4})} \\
        & IPO & 85.51 {\scriptsize (\textit{+0.83})} & 52.0 {\scriptsize (\textit{+1.0})} \\
        & \quad \textit{with weight data} & 86.58 {\scriptsize (\textit{+1.90})} & 54.0 {\scriptsize (\textit{+3.0})} \\
        \hline
        \multirow{9}{*}{\centering GLM4-9B-Chat} & Original & 58.07 & 45.2 \\
        & SFT & 61.71 {\scriptsize (\textit{+3.64})} & 42.0 {\scriptsize (\textit{-3.2})} \\
        & DPO & 61.25 {\scriptsize (\textit{+3.18})} & 46.2 {\scriptsize (\textit{+1.0})} \\
        & \quad \textit{with weight data} & 61.63 {\scriptsize (\textit{+3.56})} & \textbf{47.2} {\scriptsize (\textit{+2.0})} \\
        & DPOP & 55.40 {\scriptsize (\textit{-2.67})} & 46.2 {\scriptsize (\textit{+1.0})} \\
        & \quad \textit{with weight data} & 56.40 {\scriptsize (\textit{-1.67})} & 46.2 {\scriptsize (\textit{+1.0})} \\
        & SimPO & 58.75 {\scriptsize (\textit{+0.68})}  & 47.0 {\scriptsize (\textit{+1.8})}  \\
        & \quad \textit{with weight data} & 57.92 {\scriptsize (\textit{-0.15})}   & 46.2 {\scriptsize (\textit{+1.0})}   \\
        & IPO & 58.83 {\scriptsize (\textit{+0.76})} & 46.6 {\scriptsize (\textit{+1.4})} \\
        & \quad \textit{with weight data} & 58.68 {\scriptsize (\textit{+0.61})} & \textbf{47.2 {\scriptsize (\textit{+2.0})}} \\
        \hline
    \end{tabular}
    \label{tab:main_table}
\end{table*}

\subsection{Datasets}

Since our work involves two parts: the collection of training data and the use of test data at test time, we will elaborate on them separately.
\paragraph{Training Phase}
For training, we utilize the MetaMath~\citep{metamath} dataset, which contains 395k samples generated through data augmentation from the GSM8K~\citep{gsm8k} and MATH datasets~\cite{hendrycksmath2021}. Since the augmented data specifies its origin(GSM or MATH), we conduct separate experiments by dividing the training process into two subsets: one using data labeled as originating from MATH and the other from GSM. This separation is also reflected in the presentation of our experimental results. From the MetaMath dataset, we select the first 20k samples of MATH-type data and 20k samples of GSM-type data. These subsets were further processed through model sampling, where we set the sampling parameter \( N \) to 16, \(\alpha\) in Equation~\ref{eq:getdata} to 1 and the temperature to 0.7. A larger \( N \) allows us to better observe the model's output distribution, helping us identify questions that the model handles confidently and those where its predictions are less certain.

\paragraph{Testing Phase}
For evaluation, we employ two widely recognized datasets for assessing mathematical reasoning capabilities in models: GSM8K and MATH. The GSM8K test set consists of 1,319 math word problems, providing a benchmark for foundational mathematical reasoning. In contrast, the MATH dataset includes 5,000 problems with significantly higher difficulty levels. Due to computational constraints, we conduct our tests on a subset of the MATH dataset, referred to as MATH500~\citep{math500}. This dataset contains 500 problems and maintains the similar difficult level and subject distribution as the original MATH test set. 

\subsection{Model}
We select several series models for our experiments. Since our method requires the generation of incorrect answers, selecting models that maintain a certain error rate ensures a more comprehensive data collection process, thereby enhancing the effectiveness of training, we use ChatGLM3-6B~\citep{glm2024chatglm} for alignment with GSM data. And for MATH data, we choose Qwen2-1.5B-Instruct, Qwen2-7B-Instruct, and GLM4-9B-Chat~\citep{qwen2, glm2024chatglm}. These models have been supervised fine-tuned on common datasets, which is consistent with the typical RLHF workflow where training is conducted on a fine-tuned model rather than a pretrained one.

\subsection{Preference Optimization Baseline}
In our experiments, we compare the performance of several widely used reinforcement learning from human feedback (RLHF) methods, including DPO~\citep{dpo}, SimPO~\citep{simpo}, DPOP~\citep{dpop}, and IPO~\citep{ipo}. These methods all rely on pairwise comparisons, making it straightforward to integrate our proposed method as a plug-and-play component. Experimental results presented later demonstrate that our method consistently enhances model performance, often leading to better outcomes across various tasks.

\subsection{Training Setting}
In terms of model training, our experiments were conducted on an A100 40G GPU. Due to hardware limitations, we employed LoRA~\citep{lora} for parameter-efficient fine-tuning and used bf16 precision. Given the relatively small amount of training data, we trained models of size 1.5B and below for one epoch with a learning rate of \(2 \times 10^{-6}\). For the 7B model, we trained for half an epoch with a learning rate of \(5 \times 10^{-6}\). In our experiments, we observed that this amount of training led to a stagnation in loss reduction, with no significant improvement in performance. During training, we used the same prompts as those used during testing, which are detailed in the appendix~\ref{sec:prompt}.

\subsection{Main Result}

We report the performance of the models on GSM8K and MATH datasets, with the experimental results presented in Table \ref{tab:main_table}. The results demonstrate that the weighted method achieves a general improvement compared to methods that do not use weighted data. Additionally, we report the performance on GSM8K when the model is trained on the MATH dataset. While the improvement is not significant, it still indicates that our method is relatively stable and does not cause the trained model to overfit to a specific type of problem.

\begin{table}[t]
\centering
\begin{tabular}{l l}
\hline
\textbf{Method} & \textbf{GSM8K} \\
\hline
Original & 51.55 \\
SFT & 53.75 {\scriptsize\textit{(+2.20)}} \\
DPO & 59.96 {\scriptsize\textit{(+8.41)}} \\
\rowcolor{gray!10} \quad  \textit{with weight data} & \textbf{60.34} {\scriptsize\textit{(+8.79)}} \\
DPOP & 53.37  {\scriptsize\textit{(+1.82)}} \\
\rowcolor{gray!10} \quad \textit{with weight data} & 53.52 {\scriptsize\textit{(+1.97)}} \\
SimPO & 59.13 {\scriptsize\textit{(+7.58)}} \\
\rowcolor{gray!10} \quad  \textit{with weight data} & 59.59 {\scriptsize\textit{(+8.04)}} \\
IPO & 55.34  {\scriptsize\textit{(+3.79)}} \\
\rowcolor{gray!10} \quad \textit{with weight data} & 54.28{\scriptsize\textit{(+2.73)}} \\
\hline
\end{tabular}
\caption{The experiment result of ChatGLM3-6B on GSM8K test set among different methods.}
\label{tab:gsmresult}
\end{table}

We also conducted experiments on the GSM8K dataset, with the experimental results shown in Table~\ref{tab:gsmresult}. For the reason that the model is better to have a certain error rate we mentioned before, we selected ChatGLM3-6B as our model, which achieved slightly over 50\% accuracy on GSM8K, along with the first 20K GSM-type data from the Metamath dataset as the training data.

\subsection{Analysis}
\begin{figure}[tb]
    \centering
    \includegraphics[width=1\linewidth]{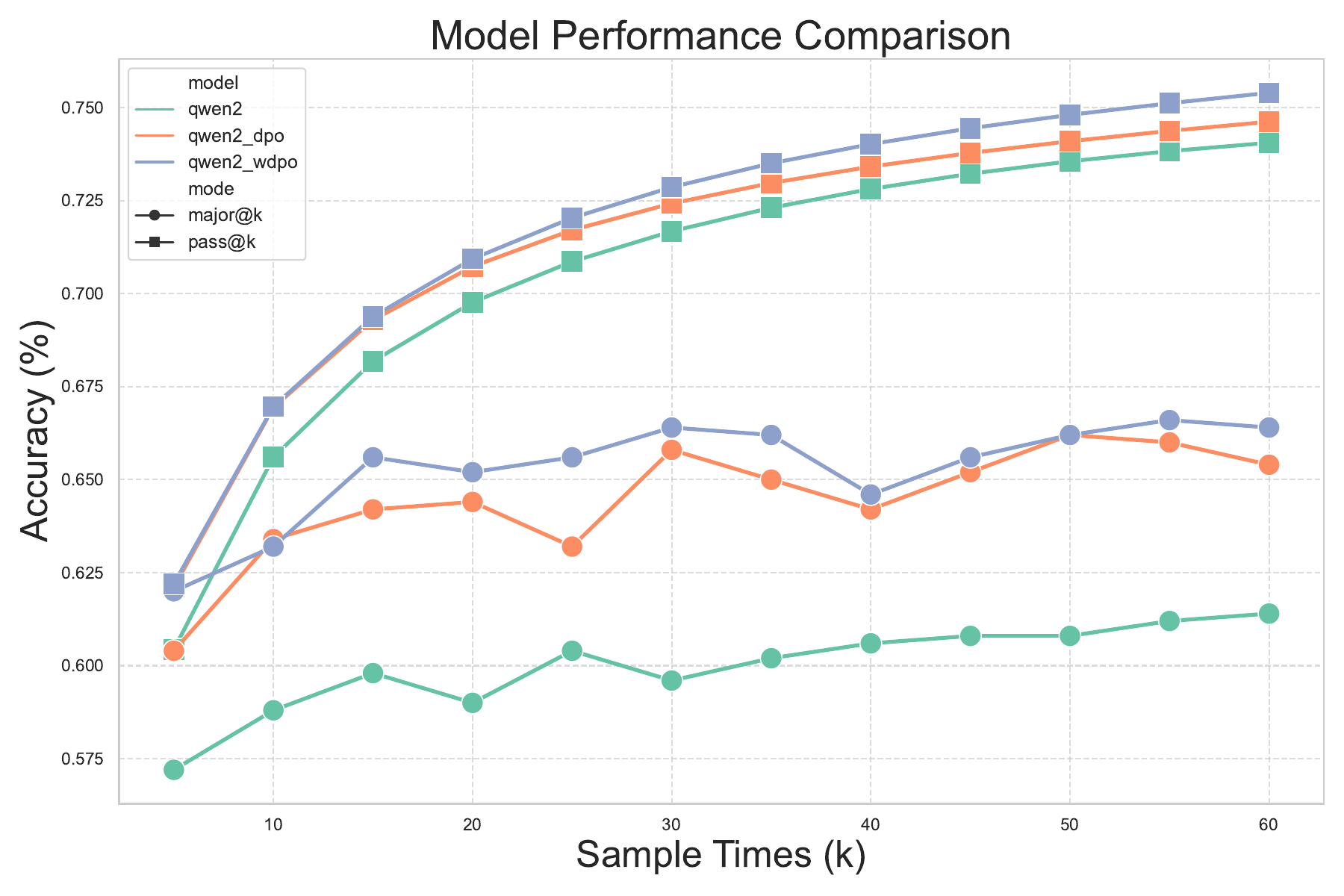}
    \caption{Results of the model evaluated with multiple samples. Specifically, we present the results for Qwen2-7B-Instruct, where it is evident that the weighted training method generally achieves higher correctness compared to the unweighted method.}
    \label{fig:majork}
\end{figure}
\paragraph{Model Stability}

To evaluate the stability of our trained model, we conduct a sampling experiment where the model generated multiple responses to the same question in MATH500. We test the stability using two methods: the major@k and the pass@k metric~\citep{majorvote, pass@k}. The experimental results are presented in Figure~\ref{fig:majork}, where we compare ordinary DPO with the weighted DPO approach. For the major@k method, we first sample \( N \) responses with temperature 0.7, then randomly selected \( k \) samples from these \( N \), and performed a majority vote on these \( k \) responses. The majority vote is calculated as the most frequent answer among the \( k \) sampled responses.

For the pass@k metric, we adopt the unbiased sampling logic described in \citet{monkeys}, using the following formula:
\begin{equation}
    \text{pass@k} = \frac{1}{|D|} \sum_{i=1}^{|D|} \left( 1 - \frac{\binom{N - C_i}{k}}{\binom{N}{k}} \right)
\end{equation}
Here,\( |D| \) represents the number of test cases and \( C_i \) represents the number of correct responses for the \( i \)-th problem. 

Since the answers to mathematical problems can be diverse, the majority vote method does not necessarily lead to higher accuracy as the number of samples increases. However, as shown in the results, models trained on weighted data generally outperform those trained on unweighted data, demonstrating the effectiveness of the proposed approach. As for pass@k, the improvement is less pronounced. We believe this is because RLHF methods like DPO primarily help the model better select the correct response rather than fundamentally enhancing its mathematical reasoning abilities.

\begin{figure}[t!]
    \centering
    \includegraphics[width=1\linewidth]{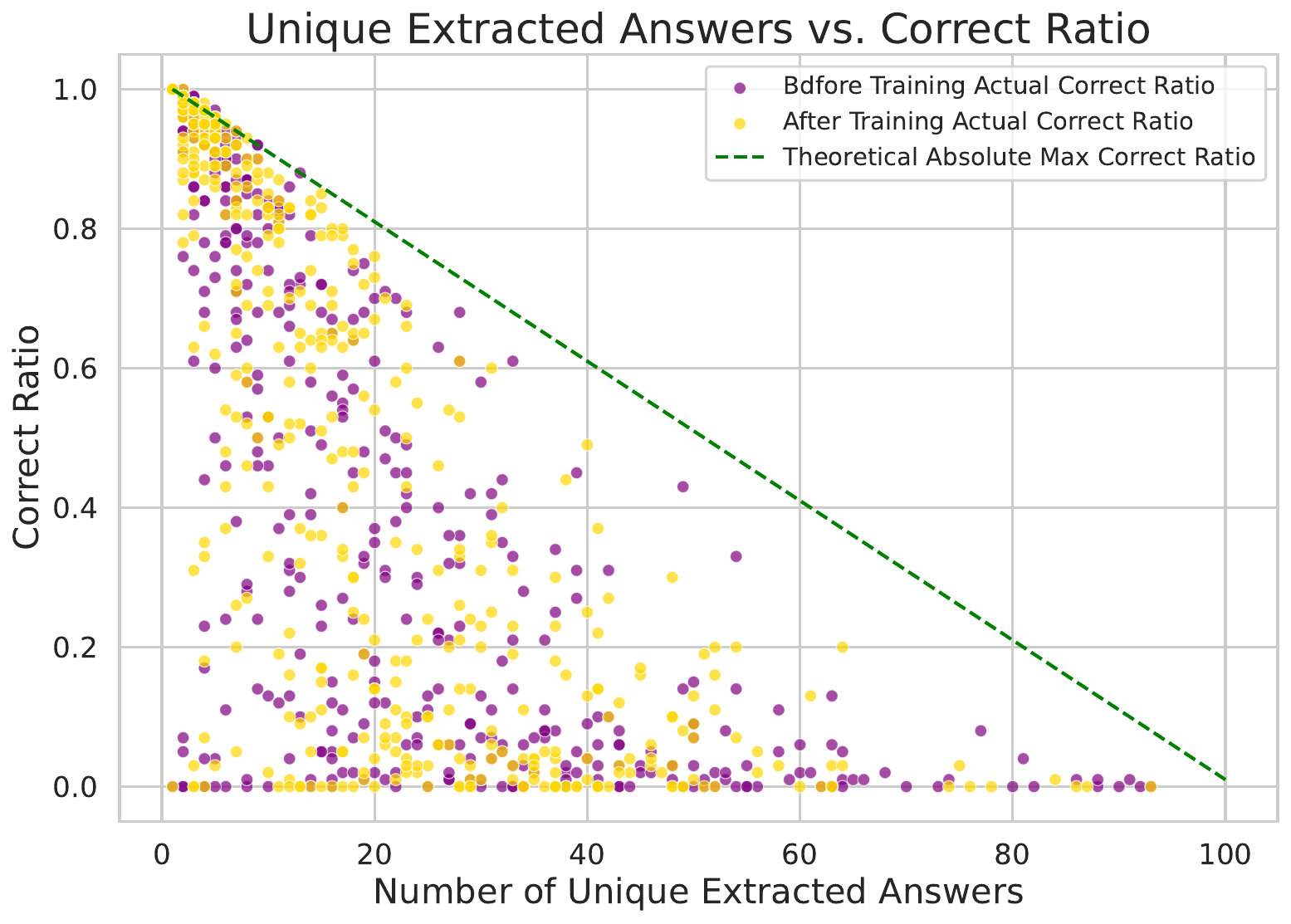}
    \caption{Data point distribution before and after training on Qwen2-7B-Instruct. The x-axis represents the number of different unique answers obtained in the sampling(which can be viewed as the number of answer equivalence classes), and the y-axis represents the proportion of correct answers among all 100 responds. The data points shift upward toward the upper left, indicating that the outputs of the weighted-trained model are more stable and accurate across multiple samples.}
    \label{fig:compare}
\end{figure}

We also compare the distribution of the model's answer after training with weighted data (as described in Section~\ref{sec:tiny}). The result is shown in Figure~\ref{fig:compare}. We observe that the data points have noticeably shifted upwards overall, indicating that the trained model is more likely to select the correct answers compared to before training.

\begin{figure}[t!]
    \centering
    \includegraphics[width=1\linewidth]{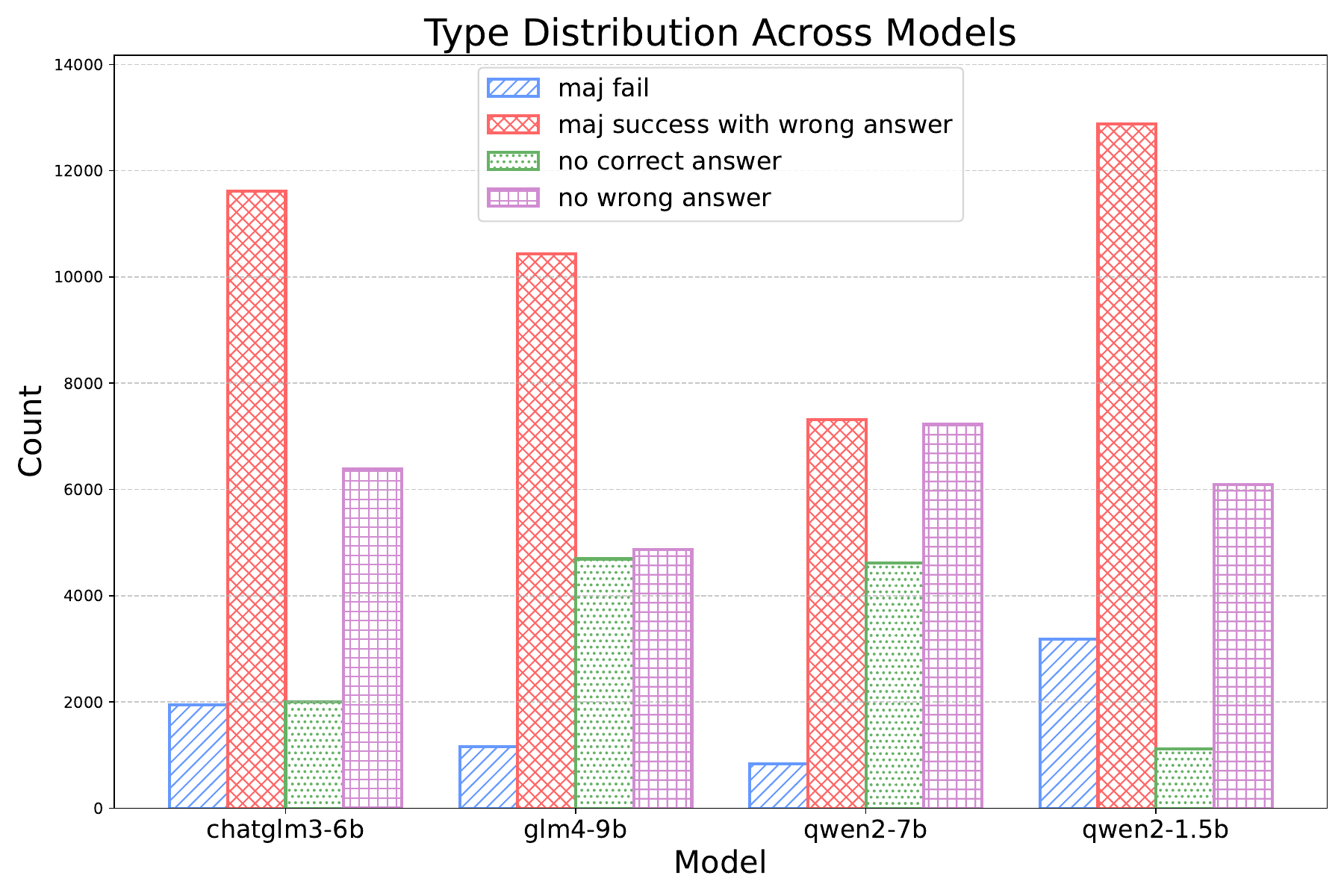}
    \caption{Training data distribution of different model.}
    \label{fig:datacount}
\end{figure}

\paragraph{Training Data Distribution}
In order to gain a deeper understanding of the training process, we analyze the distribution of the training data, and the results are shown in Figure \ref{fig:datacount}. In this analysis, ``major fail'' refer to the correct answer cannot be obtained through the majority vote method. ``major success with wrong answer'' means there are some wrong answers among the sample answers but the correct answer can be obtained through the majority vote method. ``no correct answer'' indicates that no correct answer appeared among the sampled responds. Meanwhile, ``no wrong answer'' means that there were no incorrect answers in the sample. For this subset of questions, we set the ``rejected'' response to null, meaning that these questions are not included in the training process.

\begin{figure}[t!]
    \centering
    \includegraphics[width=1\linewidth]{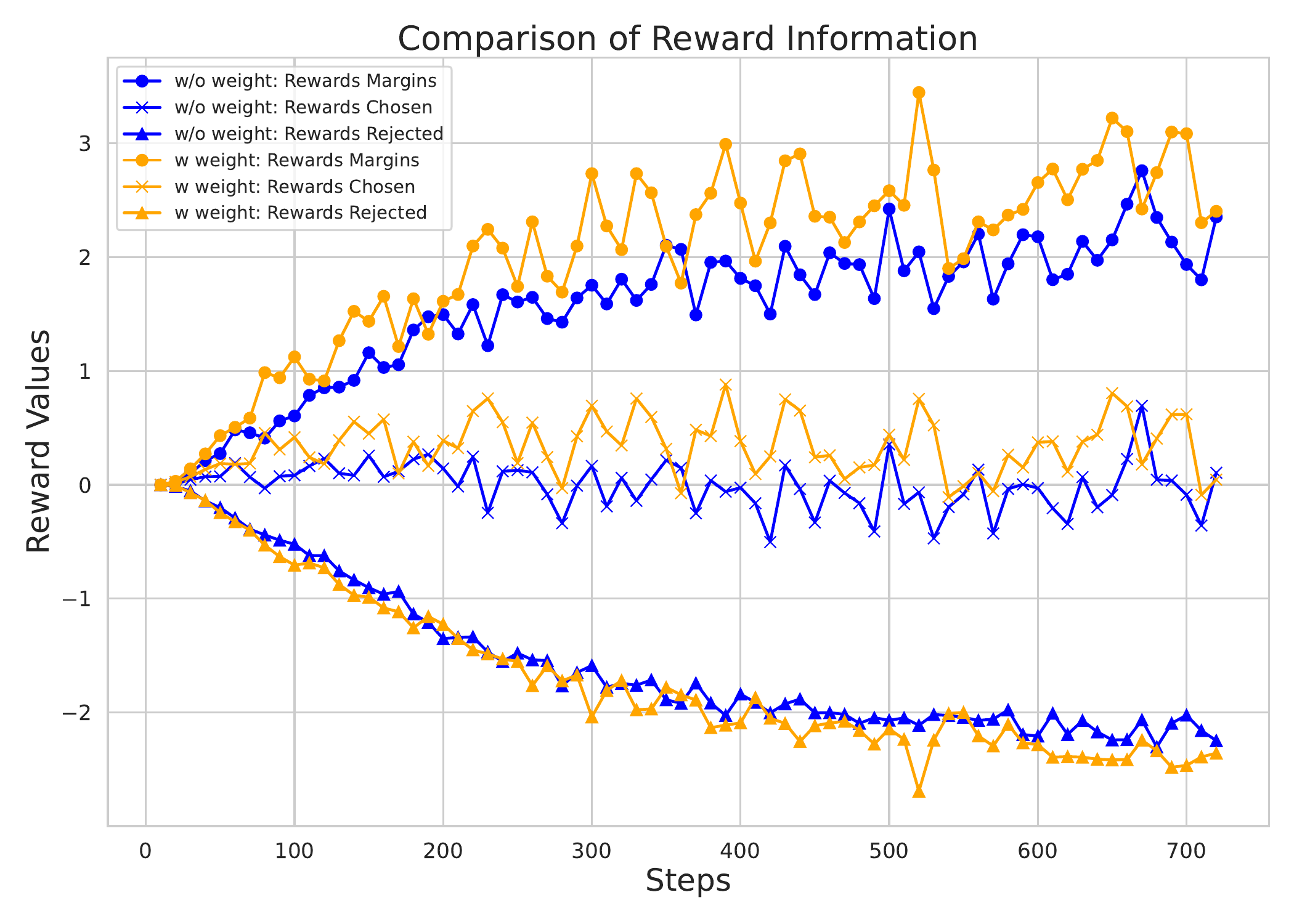}
    \caption{Reward between training with weight and without weight. Here we use DPO training on Qwen-2-7B-Instruct as example.}
    \label{fig:reward_comp}
\end{figure}

\paragraph{Reward Comparison}
To better illustrate the differences between the weighted training method and the unweighted training method during the training process, we present the results of both methods in Figure~\ref{fig:reward_comp}. The results demonstrate that the weighted training method outperforms the unweighted method in terms of both the probability of the chosen response and the rejected response. The improvement is particularly evident in the reward/chosen metric. Under the unweighted training method, the reward fluctuates around 0, whereas the weighted method consistently maintains the reward at a positive number.


\paragraph{Influence of Training Data}  
In our framework, the chosen answer is primarily the correct response generated by the model from \( N \) samples, rather than the golden answer provided by the original dataset (except in cases where the model fails to generate a correct response after multiple attempts). Selecting this self-generated content as the chosen answer helps minimize modifications to the model, thereby making \( \pi^* \) and \( \pi_{\text{ref}} \) closer to each other. In previous studies~\citep{dpop,simpo}, it has been observed that using \textbf{pairwise comparison methods} in mathematical reasoning tasks often leads to suboptimal performance. We argue that this issue arises when the golden answer is used as the chosen answer for training. The golden answer, while normative, often differs in style and format from the model's own responses. As a result, the model may focus on mimicking the \textbf{style} of the golden answer rather than learning the underlying \textbf{reasoning process}, which is critical for mathematical problem-solving.  

\begin{figure}
    \centering
    \includegraphics[width=1\linewidth]{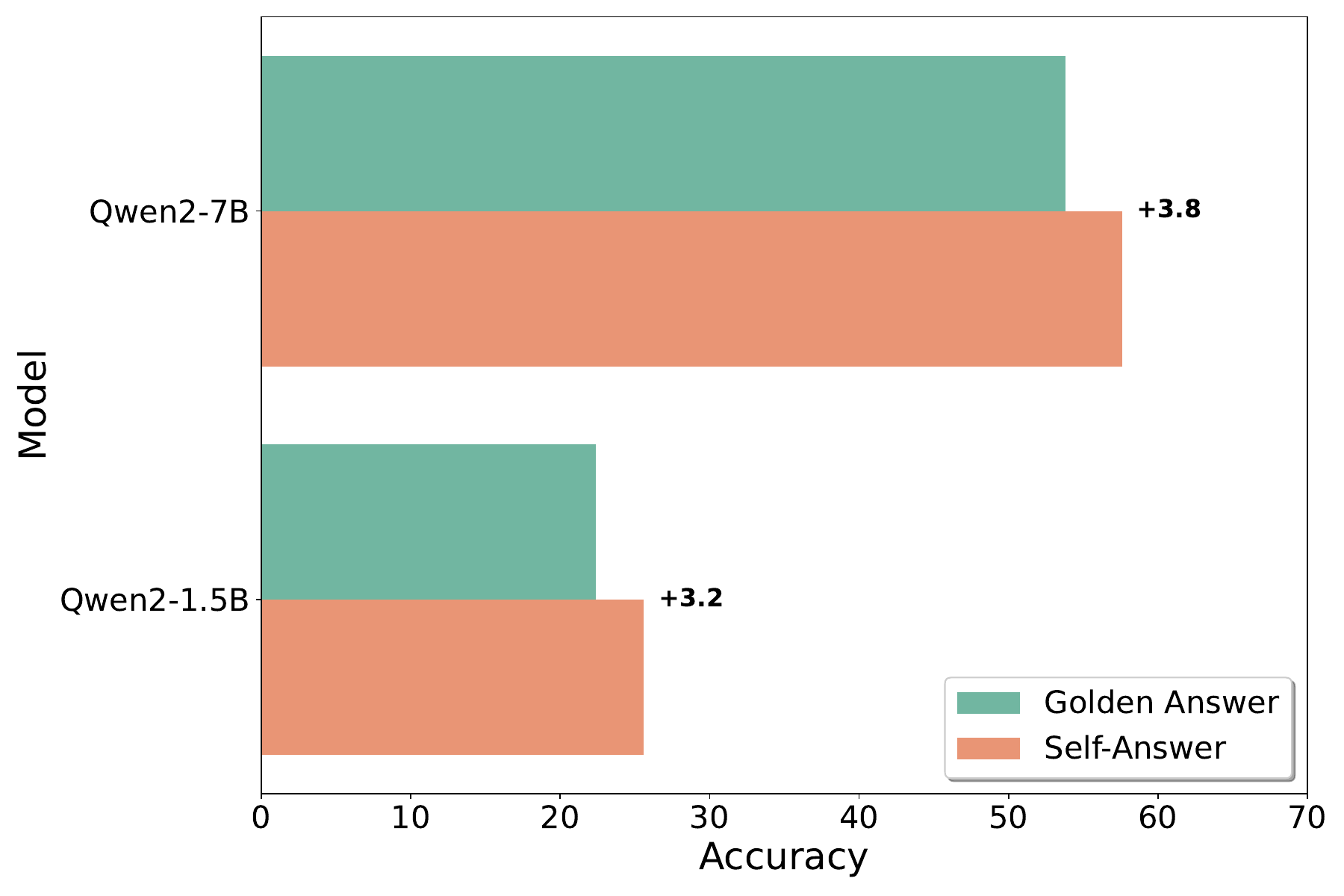}
    \caption{Comparison of Using golden Answer and Self-generated Answer}
    \label{fig:originaldata}
\end{figure}

To validate our hypothesis, we conduct a comparative experiment to evaluate the difference between using the correct response generated by the model as the chosen answer and directly using the golden answer as the chosen answer. The experimental results, shown in Figure~\ref{fig:originaldata}, indicate that our approach of using self-generated correct answers as the chosen answer outperforms the golden-answer-based approach. The results indicate that using self-generated data as the chosen answer is significantly better than using the golden answer.

\section{Conclusion}
In this work, we propose a plug-and-play weighted training framework to improve the shortcomings of the existing pairwise comparison RLHF training methods. By collecting the performance of the model on training dataset, we assign different weights to various training examples and then apply preference optimization. Our experimental results demonstrate that this framework can be successfully integrated with a range of pairwise comparison PO methods, leading to notable performance improvements in specific domains, particularly in mathematical reasoning.

\section*{Limitation}
A key limitation of our method lies in its reliance on the ability to group answers into equivalence classes. For instance, in mathematical reasoning tasks, answers with the same numeric result can be naturally treated as equivalent. However, in tasks like machine translation or reading comprehension, defining such equivalence classes becomes significantly more challenging. Some approaches, such as leveraging semantic entropy~\citep{semanticentropy} or other clustering techniques, could potentially be used to group responses into equivalence classes based on their semantic similarity or other shared attributes. By establishing a meaningful equivalence structure for responses, our method can be extended to a broader range of tasks beyond those with clearly defined correctness criteria.

\bibliography{custom}

\appendix

\section{Answer of Different Model}
\label{sec:appendixA}
\begin{figure*}
    \centering
    \includegraphics[width=1\linewidth]{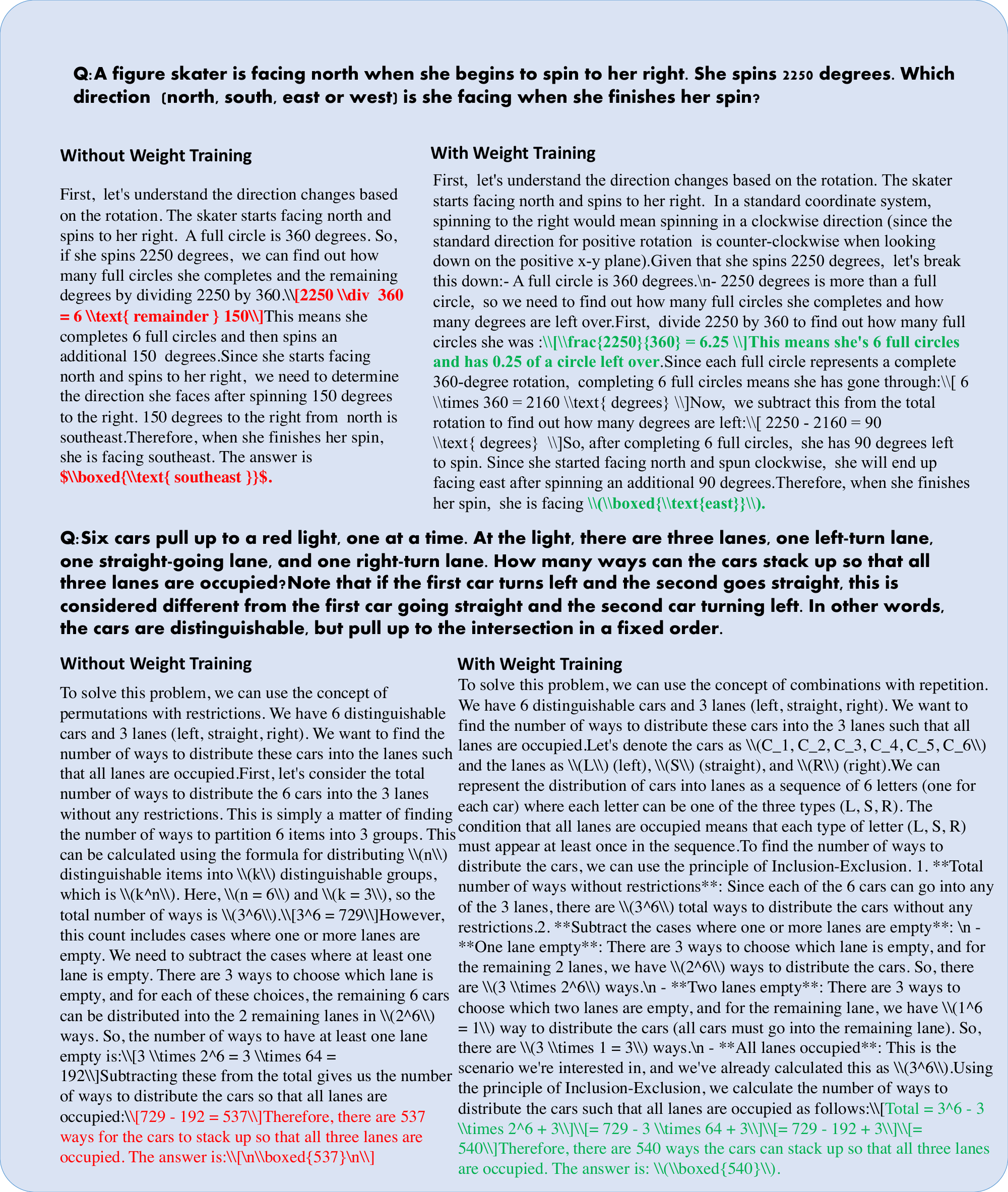}
    \caption{Comparision of examples on Qwen2-7B-Instruct with and without weighted traning}
    \label{fig:example}
\end{figure*}

In order to make a more intuitive comparison between the models obtained by weighted training and without, we choose some examples, as shown in Figure~\ref{fig:example}, to show that the answers generated by the model trained with the weighted method are correct. This comparison includes the logic, calculation steps, and the final summarized answer.

\section{The Prompt for Mathematical Problems}
\label{sec:prompt}
In our experiments, we test different prompts used by various models to solve math problems. For the Qwen2~\citep{qwen2} series models, we employ the "qwen-boxed" prompt, which is consistent with the setting used in the Step-DPO~\citep{lai2024stepdpo} method. Using this prompt, we obtain results similar to those reported in the official report.
\begin{tcolorbox}[colframe=blue!60!black, colback=blue!5, coltitle=white, title=Qwen-Boxed Prompt]
\texttt{<|im\_start|>system You are a helpful assistant.<|im\_end|>\\
<|im\_start|>user \{instruction\}\\
Please reason step by step, and put your final answer within boxed{\{\}}.<|im\_end|>\\
<|im\_start|>assistant}
\end{tcolorbox}

For the GLM4-9B-Chat~\citep{glm2024chatglm} model, we experiment with a variety of prompts, including the official chat template. However, we are unable to replicate the accuracy reported in the official results. We guess that the chat template is a general prompt for the model, and the official team might use special prompts tailored to test the model's mathematical ability. Despite this, we use a uniform prompt across all our tests, ensuring a fair comparison between the models.

As for ChatGLM3, which is primarily focused on tasks from the GSM8K dataset—where LaTeX answers in "boxed" format are uncommon—we employed a standard chain-of-thought prompt for consistency.

\begin{tcolorbox}[colframe=blue!60!black, colback=blue!5, coltitle=white, title=CoT Prompt]
\texttt{As a helpful assistant, please solve the following math problem step by step.\\
Instruction:\\
\{instruction\}
}
\end{tcolorbox}

\section{Concern about Model Selection}
In this section, we examine why our approach may not perform well on certain models, using Llama3-8B-Instruct as an example. One significant challenge is that this model does not perform well on mathematical tasks. During sampling, many questions fail to yield correct answers, forcing us to use the golden answer as the chosen answer for training. However, we observed a substantial difference in style between the golden answers from the math dataset and the answers generated by the model itself.
For example, when facing with the following problem
 The model's answer

\begin{tcolorbox}[colframe=blue!60!black, colback=blue!5, coltitle=white, title=An example on LLaMA3-8B-Instruct]
Q:A translation of the plane takes $5 + 5i$ to $6 + 8i.$  Find the complex number that the translation takes $-6$ to.

Model's Answer: To translate a complex number $z=a+bi$, we add $-a+bi$ to it. So, our translation takes $-6$ to$(-6) - 5i = -11 - 5i.$ Final Answer: The final answer is $-11-5i$. I hope it is correct.

Golden Answer: This translation takes $z$ to $z + w,$ where $w$ is a fixed complex number.  Thus, $[6 + 8i = (5 + 5i) + w.]$Hence, $w = 1 + 3i.$  Then the translation takes $-6$ to $-6 + (1 + 3i) = \boxed{-5 + 3i}.$
\end{tcolorbox}
As observed, Llama3’s response style differs significantly from the golden answer’s style. Llama3 tends to adopt a conversational tone, addressing the user directly with phrases like “we” and consistently ending responses with the sentence “I hope it is correct.” (This sentence was added to nearly all responses regardless of the question.) In contrast, the golden answer focuses solely on the reasoning process, presenting the final answer neatly formatted within a box. Despite explicitly instructing Llama3 to follow this format in the prompt, the model often fails to do so, further highlighting the discrepancy in response styles.

As a result, when training with DPO-like contrastive learning, the model tends to learn the style of the chosen answers rather than focusing on the underlying mathematical logic. To address this issue, it may be necessary to generate a set of golden answers with styles similar to the model’s responses, which can then be used as the chosen answer during training to achieve better results.
\end{document}